\documentclass{article}
\usepackage{spconf,times}
\makeatletter
\def\@sect#1#2#3#4#5#6[#7]#8{%
  \ifnum #2>\c@secnumdepth
    \let\@svsec\@empty
  \else
    \refstepcounter{#1}%
    \protected@edef\@svsec{\@seccntformat{#1}\relax}%
  \fi
  \@tempskipa #5\relax
  \ifdim \@tempskipa>\z@
    \begingroup
      #6{%
        \@hangfrom{\hskip #3\relax\@svsec}%
          \interlinepenalty \@M
          \ifnum #2=1\relax\bf\centering\uppercase{#8}\par%
          \else\ifnum #2=2\relax\bf #8\par%
          \else\it #8\par\fi\fi}%
    \endgroup
    \csname #1mark\endcsname{#7}%
    \addcontentsline{toc}{#1}{%
      \ifnum #2>\c@secnumdepth \else
        \protect\numberline{\csname the#1\endcsname}%
      \fi
      #7}%
  \else
    \def\@svsechd{%
      #6{\hskip #3\relax
      \@svsec \ifnum #2=1\relax\bf\uppercase{#8}\else#8\fi}%
      \csname #1mark\endcsname{#7}%
      \addcontentsline{toc}{#1}{%
        \ifnum #2>\c@secnumdepth \else
          \protect\numberline{\csname the#1\endcsname}%
        \fi
        #7}}%
  \fi
  \@xsect{#5}}
\makeatother

\usepackage[utf8]{inputenc}
\usepackage[T1]{fontenc}
\usepackage{amsmath,amssymb}
\usepackage{graphicx}
\usepackage{booktabs}
\usepackage{multirow}
\usepackage{array}
\usepackage{url}
\usepackage{hyperref}
\usepackage{bm}
\usepackage{xcolor}

\title{PatTransBench: An IPC-Stratified, Multi-Dimensional Benchmark for Patent Machine Translation Evaluation}

\name{Xiaowei Su$^{1,2}$ \qquad Xusen Cheng$^{1,*}$}
\address{$^{1}$Renmin University of China, Beijing, China \\
  $^{2}$Beijing Bayuegua Technology Co., Ltd., Beijing, China \\
  \{xwsu, xusen.cheng\}@ruc.edu.cn \\
  $^{*}$Corresponding author \\
  ORCID: 0000-0002-7784-7141 (Su), 0000-0002-2028-1337 (Cheng)}

\begin{document}
\ninept
\maketitle
\setlength{\textfloatsep}{6pt plus 1pt minus 1pt}
\setlength{\floatsep}{6pt plus 1pt minus 1pt}
\setlength{\abovecaptionskip}{4pt}
\setlength{\belowcaptionskip}{2pt}

\newcommand{\lsep}{\slash\allowbreak}
\newcommand{\mpnet}{paraphrase-\allowbreak multilingual-\allowbreak mpnet-\allowbreak base-v2}
\begin{abstract}
We present PatTransBench, a 12,000-item Chinese--English\lsep Japanese\lsep Korean patent machine translation (PatMT) benchmark with a four-dimensional evaluation protocol. The benchmark comprises 6,000 short-text (sentence/claim) and 6,000 long-text (paragraph) pairs, evenly split across three language pairs and both translation directions, built via IPC-section-stratified sampling and dual-threshold quality gating from upstream patent-family parallel corpora. The evaluation protocol spans literal $n$-gram overlap (BLEU, ROUGE-L, chrF), cross-lingual semantic textual similarity (CL-STS), 38-probe-language consistency testing, and serving performance under concurrency. A case study on PatMT2-7B and its fp8/AWQ4/AutoRound-4bit variants reveals direction-dependent quantization effects masked by aggregate scores and non-monotonic language-failure patterns across quantization schemes that are invisible to single-metric evaluation. We plan to release the benchmark, protocol, and case-study results upon publication to support reproducible PatMT evaluation.
\end{abstract}


\section{Introduction}
\label{sec:intro}

Patent claims and specifications are written in dense, legally load-bearing technical prose where a mistranslated hyphen or a dropped negation can change the scope of a legal right, and the same document must often be produced consistently across Chinese, English, Japanese, and Korean for multi-jurisdictional filing. We refer to this task throughout as Patent Machine Translation (PatMT). Despite its practical weight, most reported evaluations of patent-oriented PatMT systems still rely on a single corpus-level BLEU or ROUGE score computed on whatever internal test split happened to be available. A single scalar cannot distinguish a model that paraphrases correctly from one that has quietly started answering in the wrong language, nor can it tell a practitioner whether a quantized deployment variant will hold up under concurrent production load.

PatTransBench addresses these gaps through three design decisions: (1)~\emph{IPC-stratified sampling} with honest reporting when stratification fails due to upstream data scarcity; (2)~\emph{precise metric naming}, using Cross-lingual Semantic Textual Similarity (CL-STS) computed from a named multilingual sentence-embedding model; (3)~\emph{multi-dimensional evaluation} that surfaces findings no single metric reveals. The PatTransBench dataset is currently being prepared for public release and will be made available at https://huggingface.co/datasets/SuStarSmile/PatTransBench. Our contributions are:
\begin{itemize}
\item A 12,000-item IPC-stratified PatMT benchmark with dual-threshold quality gating, covering three language pairs with honest reporting of stratification outcomes.
\item A four-dimensional evaluation protocol spanning literal overlap, CL-STS, language consistency, and serving performance.
\item A case study revealing direction-dependent quantization effects and non-monotonic language-failure patterns across quantization schemes that no single dimension would surface.
\end{itemize}

\begin{figure}[t]
\centering
\includegraphics[width=\columnwidth]{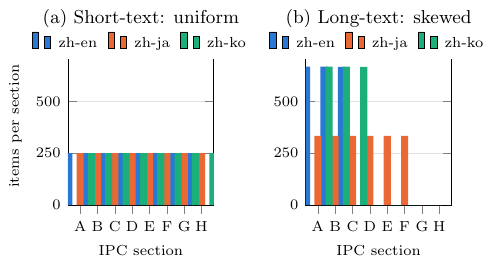}
\caption{Per-IPC-section item counts. Short-text (left): exact 250-item uniform coverage across all eight sections. Long-text (right): skewed to A--C (zh-en/zh-ko) and A--F (zh-ja) after dual-threshold quality gating.}
\label{fig:ipc-dist}
\end{figure}

Existing MT benchmarks such as WMT shared tasks and FLORES~\cite{nllb2024scaling} provide broad-coverage evaluation sets drawn from news and Wikipedia, but are not stratified by technical domain and do not evaluate serving performance or output-language consistency under fine-tuning. Standard MT metrics including BLEU~\cite{papineni2002bleu}, ROUGE~\cite{lin2004rouge}, and chrF~\cite{popovic2015chrf} capture surface-level overlap but miss semantic fidelity and deployment viability. Learned metrics such as COMET~\cite{rei2020comet} require human-annotated training data; our CL-STS instead scores semantic agreement directly with a fixed multilingual sentence encoder, at no annotation cost and with full reproducibility.

Evaluating a benchmark of this scope also requires a defensible answer to a practical question: on what evidence should a deployment decision rest? We build the four dimensions so that each targets a failure mode the others cannot detect on their own, and we report the disagreements among them rather than averaging them into a single number. The case study is a demonstration of that principle on one model family and its quantization variants, not a cross-system comparison; construction, coverage statistics, and protocol stand independently of which systems are ultimately scored against them.

\section{Benchmark Construction}
\label{sec:construction}

\subsection{Composition and IPC Stratification}

PatTransBench contains 12,000 items: 6,000 short-text (sentence- and claim-level pairs) and 6,000 long-text (paragraph-level pairs drawn from full patent specifications), evenly split across three language pairs---Chinese-English (zh-en), Chinese-Japanese (zh-ja), Chinese-Korean (zh-ko)---at 2,000 items per pair. Each item carries both translation directions, so counts refer to aligned pairs rather than single-direction instructions.

Both pipelines use IPC-stratified sampling targeting eight sections (A through H). Given a per-language-pair budget $N=2000$ and section set $\mathcal{S}$ with $|\mathcal{S}|=8$, the per-section target is:
\begin{equation}
\label{eq:ipc-limit}
n_s = \left\lfloor \frac{N}{|\mathcal{S}|} \right\rfloor + \mathbb{1}\left[\text{rank}(s) < N \bmod |\mathcal{S}|\right] = 250
\end{equation}
When a section returns fewer rows than requested, the deficit is redistributed across sections with non-zero supply using the same formula, fetched at an offset past already-retrieved rows. If a second-pass deficit remains, it is logged as a warning. Short-text achieves exact uniform coverage: all 24 first-pass queries returned exactly 250 rows, confirmed in extraction logs. Long-text is skewed: although upstream patent-family data covers all eight IPC sections for every language pair, the dual-threshold quality gate (Section~\ref{sec:quality-gate}) filters out a substantial fraction of candidates, and the remaining qualified items are unevenly distributed across sections. After quality gating and top-up rebalancing, zh-en and zh-ko cover only sections A--C (667/667/666), zh-ja covers A--F (334/334/333/333/333/333), with zero items in remaining sections. This reflects the interaction between patent-family density and quality-gate stringency across IPC domains, not a sampling defect. Figure~\ref{fig:ipc-dist} visualizes both outcomes.

\subsection{Dual-Threshold Quality Gating}
\label{sec:quality-gate}

Short-text candidates require $\text{similarity\_score} > 0.9$. Long-text candidates must satisfy a conjunction of two independent embedding-model thresholds plus a language-pair-specific word-count-ratio range:
\begin{equation}
\label{eq:long-gate}
\text{sim}_{\text{0.6B}} > 0.9 \;\land\; \text{sim}_{\text{8B}} > 0.8 \;\land\; r_{\min} < \tfrac{|\text{words(tgt)}|}{|\text{words(src)}|} < r_{\max}
\end{equation}
where $\text{sim}_{\text{0.6B}}$ and $\text{sim}_{\text{8B}}$ are precomputed scores from a LoRA-adapted 0.6B and an 8B embedding model respectively; requiring agreement from two independently-sized models is stricter than either alone. Word-ratio ranges are $(0.5, 0.9)$ for zh-en, $(1.0, 1.5)$ for zh-ja, and $(1.0, 1.6)$ for zh-ko, reflecting different average expansion/contraction ratios. Both gates are applied jointly with IPC filtering in a single SQL query.

\section{Evaluation Protocol}
\label{sec:protocol}

\textbf{Dimension 1: Literal Overlap.} We report corpus-level BLEU via sacrebleu~\cite{post2018sacrebleu}, ROUGE-L~\cite{lin2004rouge}, and chrF in chrF2 configuration (character order 6, word order 0, $\beta=2$)~\cite{popovic2015chrf,popovic2017chrfpp}. For Japanese and Korean targets we use MeCab-based tokenizers rather than whitespace segmentation, since whitespace tokenization is not meaningful for these languages.

\textbf{Dimension 2: Cross-lingual Semantic Textual Similarity.} We compute CL-STS as cosine similarity of L2-normalized [CLS] embeddings from \mpnet~\cite{reimers2020multilingual}:
\begin{equation}
\label{eq:cl-sts}
\text{CL-STS}(h, r) = \frac{\mathbf{e}_h \cdot \mathbf{e}_r}{\|\mathbf{e}_h\|_2 \, \|\mathbf{e}_r\|_2}, \quad \mathbf{e} = \text{CLS}(\text{Enc}(\cdot))
\end{equation}
{\emergencystretch=2em
The encoder is the multilingual sentence-embedding model \mpnet, an XLM-RoBERTa network distilled across 50+ languages so that equivalent sentences in different languages land close together in one shared embedding space. We score both hypothesis and reference with this frozen encoder, so the resulting number measures geometric agreement between two vectors rather than a quality estimate produced by a trained regression head; it needs no annotated data and is exactly reproducible from the released script and model checkpoint.
\par}

\textbf{Dimension 3: Language Consistency.} We probe 38 target languages covering major world languages, low-resource and non-Latin-script languages, and script variants (e.g., Traditional Chinese, Cantonese). For each, we issue a fixed-source translation instruction and inspect raw outputs for two failure modes: repetition-collapse (a short $n$-gram repeated far beyond fluent output) and instruction-echo (output translates the instruction rather than the source). Automated non-empty-response checks are insufficient; only direct text inspection reveals these failures.

\textbf{Dimension 4: Serving Performance.} We measure TTFT (avg/$p$95/$p$99), TPOT, QPS, and TPS under concurrent load at levels 10/20/50, separately for short and long text, since long documents produce qualitatively different latency and throughput profiles.

\section{Case Study: PatMT2}
\label{sec:case-study}

We apply the full protocol to PatMT2-7B, a bilingual/multilingual PatMT model fine-tuned in two stages on zh-en/ja/kr patent data, and three quantized variants: fp8, AWQ 4-bit (awq4), AutoRound 4-bit (auto4). This is a demonstration on one model family rather than a cross-system comparison: we hold no independently re-verified numbers for other systems under one frozen test set and matched decoding, so we present the within-family disagreements the four dimensions expose and leave cross-system ranking to the released benchmark.

\subsection{Direction-Dependent Quantization Effects}

Table~\ref{tab:quality} reports BLEU/ROUGE-L/chrF across all six directions. Quantization does \emph{not} degrade quality uniformly: on ja$\to$zh, fp8 drops 4.4 BLEU points below base (48.38$\to$43.95) while awq4 stays within 0.1 points; auto4 lands between them. The same direction shows the largest ROUGE-L/chrF spread across variants. This direction-specific effect is masked by cross-direction averaging---a practitioner choosing between fp8 and awq4 from aggregate scores alone would not see it. CL-STS tracks literal-overlap metrics directionally, confirming broad consistency at this quantization stage.

\begin{table}[t]
\centering
\caption{BLEU / ROUGE-L / chrF by variant and direction (4,000 items/direction).}
\label{tab:quality}
\small
\begin{tabular}{llccc}
\toprule
Model & Dir. & BLEU & ROUGE-L & chrF \\
\midrule
\multirow{6}{*}{base}
 & zh$\to$en & \textbf{40.31} & 72.70 & \textbf{68.04} \\
 & en$\to$zh & 42.96 & 54.31 & \textbf{42.21} \\
 & zh$\to$ja & 48.65 & 13.02 & \textbf{53.34} \\
 & ja$\to$zh & 48.38 & 39.63 & \textbf{49.03} \\
 & zh$\to$ko & 36.44 & \textbf{42.48} & 44.56 \\
 & ko$\to$zh & 43.59 & 40.54 & \textbf{41.49} \\
\midrule
\multirow{6}{*}{fp8}
 & zh$\to$en & 40.18 & \textbf{72.71} & 67.93 \\
 & en$\to$zh & 42.70 & \textbf{54.38} & 42.13 \\
 & zh$\to$ja & \textbf{48.69} & \textbf{13.20} & 53.17 \\
 & ja$\to$zh & 43.95 & 39.53 & 48.19 \\
 & zh$\to$ko & 36.11 & 42.35 & 44.25 \\
 & ko$\to$zh & 43.43 & \textbf{40.75} & 41.02 \\
\midrule
\multirow{6}{*}{awq4}
 & zh$\to$en & 40.10 & 72.57 & 68.01 \\
 & en$\to$zh & \textbf{43.64} & 54.03 & 41.77 \\
 & zh$\to$ja & 48.65 & 12.67 & 53.09 \\
 & ja$\to$zh & \textbf{48.45} & \textbf{40.27} & 48.40 \\
 & zh$\to$ko & \textbf{36.73} & 42.35 & \textbf{44.62} \\
 & ko$\to$zh & \textbf{43.86} & 40.65 & 40.97 \\
\midrule
\multirow{6}{*}{auto4}
 & zh$\to$en & 39.71 & 72.46 & 67.85 \\
 & en$\to$zh & 43.10 & 54.09 & 42.02 \\
 & zh$\to$ja & 47.76 & 12.95 & 52.42 \\
 & ja$\to$zh & 45.88 & 39.78 & 48.23 \\
 & zh$\to$ko & 36.41 & 42.31 & 44.41 \\
 & ko$\to$zh & 43.29 & 40.65 & 40.80 \\
\bottomrule
\end{tabular}
\end{table}

\subsection{Persistent Four-Language Failure Mode}

34 of 38 probed languages produce fluent, correctly-scripted output under the BF16 and fp8 variants. Four languages fail under BF16/fp8 (Table~\ref{tab:lang-fail}): Tibetan, Mongolian, and Tamil exhibit repetition-collapse; Uyghur echoes the instruction. The failure set varies across quantization schemes: AutoRound-4bit recovers three of the four (Tamil, Mongolian, Uyghur) but introduces a new failure on Gujarati (37/38); AWQ-4bit recovers Tamil and Mongolian but fails on Tibetan and Uyghur (36/38). That the core weakness concentrates in low-resource, non-CJK, non-Latin-adjacent scripts across all variants suggests a systematic representational fragility rather than a sampling artifact. Automated non-empty-response checks report 38/38 success for every variant; only direct text inspection reveals these failures.

\begin{table}[t]
\centering
\caption{Language-consistency failures (BF16/fp8, representative). Failure sets vary across quantization variants; see text for details.}
\label{tab:lang-fail}
\small
\begin{tabular}{lccp{3.8cm}}
\toprule
Lang. & Chars & Time(s) & Failure \\
\midrule
Tibetan & 18,140 & 177.4 & Phrase repeated 464$\times$ \\
Mongolian & 9,798 & 177.4 & Token loop \\
Tamil & 14,020 & 137.6 & ``Moreover'' repeated 1,949$\times$ \\
Uyghur & 45 & 0.3 & Instruction echo \\
\bottomrule
\end{tabular}
\end{table}

\subsection{Serving Performance Trade-offs}

\begin{figure}[t]
\centering
\includegraphics[width=\columnwidth]{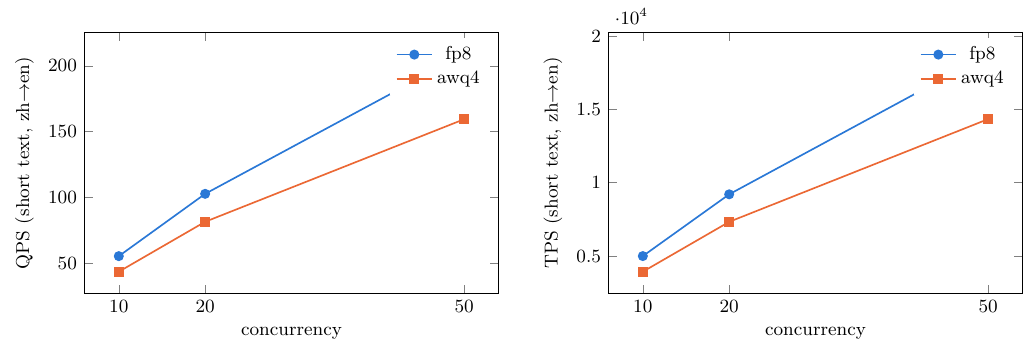}
\caption{QPS and TPS vs.\ concurrency, zh$\to$en short-text, fp8 vs.\ awq4. Both scale near-linearly; fp8 leads at every concurrency level.}
\label{fig:throughput}
\end{figure}

Table~\ref{tab:serving} reports serving metrics for zh$\to$en. fp8 outperforms awq4 by 25--30\% on QPS at every concurrency level for short-text. Long-text QPS is $\sim$2 orders of magnitude lower than short-text, but TPS is much closer between data types because TPS aggregates over many more output tokens per request. Figure~\ref{fig:throughput} shows near-linear scaling with concurrency. Quality tables alone would not predict this throughput ordering; deployment decisions require both quality and performance dimensions.

\begin{table}[t]
\centering
\caption{Serving performance, zh$\to$en, fp8 vs.\ awq4 (2,000 items/cell).}
\label{tab:serving}
\small
\begin{tabular}{llcccc}
\toprule
Var. & Type & Conc. & QPS & TPS & TTFT avg \\
\midrule
\multirow{6}{*}{fp8}
 & short & 10 & 55.47 & 4995 & 0.018s \\
 & short & 20 & 102.79 & 9207 & 0.019s \\
 & short & 50 & 208.79 & 18760 & 0.023s \\
 & long & 10 & 1.97 & 3966 & 0.073s \\
 & long & 20 & 3.02 & 6115 & 0.079s \\
 & long & 50 & 4.51 & 9077 & 0.101s \\
\midrule
\multirow{6}{*}{awq4}
 & short & 10 & 43.66 & 3929 & 0.024s \\
 & short & 20 & 81.57 & 7331 & 0.023s \\
 & short & 50 & 159.47 & 14339 & 0.030s \\
 & long & 10 & 1.60 & 3253 & 0.117s \\
 & long & 20 & 2.55 & 5128 & 0.125s \\
 & long & 50 & 4.03 & 8138 & 0.155s \\
\bottomrule
\end{tabular}
\end{table}

\subsection{What the Four Dimensions Reveal Together}

Taken individually, each dimension supports a plausible but incomplete recommendation. Quality scores alone suggest awq4 is interchangeable with fp8. Serving numbers alone suggest fp8 wins with no quality context. Language-consistency testing alone suggests all variants share the same low-resource language weakness. Only together do we get the actionable picture: fp8 is faster and not meaningfully worse on five of six directions, but meaningfully worse on ja$\to$zh; quantization can both recover and introduce language failures (AutoRound recovers 3/4 BF16 failures but adds Gujarati); and no single variant dominates all dimensions. This is the central argument for multi-dimensional evaluation: the failure modes that matter for deployment do not all appear in the same metric.

\section{Conclusion}
\label{sec:conclusion}

PatTransBench provides a 12,000-item IPC-stratified patent MT benchmark with a four-dimensional evaluation protocol. Our case study demonstrates that multi-dimensional evaluation surfaces actionable findings---direction-dependent quantization effects and non-monotonic language-failure patterns across quantization schemes---that no single metric reveals. Limitations include: (1)~skewed long-text IPC coverage (A--C/A--F only, due to quality-gate filtering reducing qualified candidates in other sections despite upstream data availability); (2)~absence of human evaluation (planned with $\sim$260 items, 3 annotators/language, adequacy+fluency Likert scales); (3)~CL-STS correlation with human judgments not yet validated; (4)~no systematic quantization ablation beyond four variants; (5)~case study limited to one model family. Future work includes expanding long-text IPC coverage, validating CL-STS against human judgments and trained COMET, extending the protocol to distilled/pruned models, and incorporating diverse translation systems for cross-system comparison. We plan to release all code, data, and results upon publication.

\bibliographystyle{IEEEbib}
\bibliography{references}

\appendix
\section{The 38 Probe Languages}
\label{app:languages}

Chinese (zh), English (en), French (fr), Portuguese (pt), Spanish (es), Japanese (ja), Turkish (tr), Russian (ru), Arabic (ar), Korean (ko), Thai (th), Italian (it), German (de), Vietnamese (vi), Malay (ms), Indonesian (id), Filipino/Tagalog (tl), Hindi (hi), Traditional Chinese (zh-TW), Polish (pl), Czech (cs), Dutch (nl), Khmer (km), Burmese (my), Persian (fa), Gujarati (gu), Urdu (ur), Telugu (te), Marathi (mr), Hebrew (he), Bengali (bn), Tamil (ta), Ukrainian (uk), Tibetan (bo), Kazakh (kk), Mongolian (mn), Uyghur (ug), Cantonese (yue). 34 of these produce fluent, correctly scripted output on direct inspection under the base BF16 variant; Tibetan, Mongolian, Tamil, and Uyghur fail as detailed in Table~\ref{tab:lang-fail}.

\end{document}